\documentclass[sigconf]{acmart}
\AtBeginDocument{%
  \providecommand\BibTeX{{%
    \normalfont B\kern-0.5em{\scshape i\kern-0.25em b}\kern-0.8em\TeX}}}

\acmConference[Archive 2021]{Archive}{August 08, 2021}{.}
\acmBooktitle{Archive ’21}

\begin{document}

\title{Efficacy of BERT embeddings on predicting disaster from Twitter data}

\author{Ashis Kumar Chanda}

\email{ashis@temple.edu}
\affiliation{
  \institution{Temple University}
  \city{Philadelphia}
  \state{PA}
  \country{USA}
}

\renewcommand{\shortauthors}{Chanda. et al.}

\begin{abstract}
Social media like Twitter provide a common platform to share and communicate personal experiences with other people. People often post their life experiences, local news, and events on social media to inform others. Many rescue agencies monitor this type of data regularly to identify disasters and reduce the risk of lives. However, it is impossible for humans to manually check the mass amount of data and identify disasters in real-time. For this purpose, many research works have been proposed to present words in machine-understandable representations and apply machine learning methods on the word representations to identify the sentiment of a text. The previous research methods provide a single representation or embedding of a word from a given document. However, the recent advanced contextual embedding method (BERT) constructs different vectors for the same word in different contexts. BERT embeddings have been successfully used in different natural language processing (NLP) tasks, yet there is no concrete analysis of how these representations are helpful in disaster-type tweet analysis. In this research work, we explore the efficacy of BERT embeddings on predicting disaster from Twitter data and compare these to traditional context-free word embedding methods (GloVe, Skip-gram, and FastText). We use both traditional machine learning methods and deep learning methods for this purpose. We provide both quantitative and qualitative results for this study. The results show that the BERT embeddings have the best results in disaster prediction task than the traditional word embeddings. Our codes are made freely accessible to the research community.

\end{abstract}

\keywords{Twitter data, social media data, disaster prediction, BERT, Kaggle competition, natural language processing (NLP)}

\maketitle

\section{Introduction}
In the current age of internet, online social media sites have become available to all people, and people tend to post their personal experiences, current events, local and global news. For this reason, the daily usages of social media are growing up and making a large dataset that becomes an important source of data to make different types of research analysis. Moreover, the social media data are real time data and accessible to monitor. Therefore, several research works are conducted to perform different types real time predictions using the social media data, such as stock movement prediction \cite{nguyen2015sentiment}, relation extraction \cite{ritter2015weakly}, and natural disaster prediction \cite{yoo2018social,leung2020big}.

Twitter is such a social media site that can be accessed through people's laptops and smartphones. The rapid growth of smartphone or laptop usages enables people to share an emergency that they observe in real time. For this reason, many disaster relief organizations and news agencies are interested in monitoring Twitter data programmatically. However, unlike long articles, tweets are short length text, and they tend to have more challenges due to their shortness, sparsity (i.e., diverse word content) \cite{chen2011short}, velocity (rapid growth of short text like SMS and tweet) and misspelling \cite{alsmadi2019review}. For these reasons, it is very challenging to understand whether a person's words are announcing a disaster or not. For example, a tweet like this, "$\#oldBand$ $amazing$ $performance$! $light,$ $color,$ $fire$ $on$ $stage$! $lots$ $of$ $people$ $and$ $huge$ $chaos!$" tells us an experience of a person in a concert and we can say from that he enjoyed it, because of the word, "$amazing$". Even though it contains the word, "$fire$", it does not mean any danger or emergency; rather, it is used to describe the colorful decoration of the stage. Let's assume another tweet like this, "$California$ $Hwy.$ $20$ $closed$ $in$ $both$ $directions$ $due$ $to$ $Lake$ $Country$ $fire$". Here, the word ``$fire$" means disaster, and the tweet describes an emergency. The two examples show that one word could have multiple meanings based on its context. Therefore, understanding the context of words is important to analyze a tweet's sentiment.

Different researchers proposed different methods to understand the meaning of a word by representing it in embedding or vector \cite{pennington2014glove,DBLP:journals/corr/abs-1301-3781,bojanowski2016enriching}. Neural network-based methods such as Skip-gram \cite{DBLP:journals/corr/abs-1301-3781}, FastText \cite{bojanowski2016enriching} are popular for learning word embeddings from large word corpus and have been used for solving different types of NLP tasks. These methods are also used for sentiment analysis of Twitter data \cite{deho2018sentiment,poornima2020comparative}. However, those embedding learning methods provide static embedding for a single word in a document. Hence, the meaning of the word,``$fire$" would remain the same in the above two examples for these methods.

To handle this problem, the authors of \cite{devlin2018bert} proposed a contextual embedding learning model, Bidirectional Encoder Representations from Transformers (BERT), that provides embeddings of a word based on its context words. In different types of NLP tasks such as text classification \cite{sun2019fine}, text summarization \cite{liu2019text}, entity recognition \cite{hakala2019biomedical}, BERT model outperformed traditional embedding learning models. However, it is interesting to discover how the contextual embeddings could help to understand disaster-type texts. For this reason, we plan to analyze the disaster prediction task from Twitter data using both context-free and contextual embeddings in this study. We use traditional machine learning methods and neural network models for the prediction task where the word embeddings are used as input to the models. We show that contextual embeddings work better in predicting disaster-types tweets than the other word embeddings. Finally, we provide an extensive discussion to analyze the results.

The main contributions of this paper are summarized as follows.
\begin{enumerate}
    \item We analyze a real-life natural language online social network dataset, Twitter data, to identify challenges in human sentiment analysis for disaster-type tweet prediction.
    
    \item We apply both contextual and context-free embeddings in tweet representations for disaster prediction through machine learning methods and show that context-free embeddings (BERT) can improve the accuracy of disaster prediction compared with contextual embeddings.
    
    \item We provide a detailed explanation of our method and results and share our codes publicly that will enable researchers to run our experiments and reproduce our results for future research work directions \footnote{\label{our_code} https://github.com/ashischanda/sentiment-analysis}. 
\end{enumerate}


The rest of the paper is organized as follows. In section 2, some related works are introduced. The main methodology of this paper is elaborated in section 3. The dataset and the experiments are presented in section 4 and 5, respectively. Finally, the conclusion is drawn in section 6.

\section{Related Works}
Many research works analyzed Twitter data for understanding emergency situation and predicting disaster analysis \cite{karami2020twitter, zou2018mining, ashktorab2014tweedr, olteanu2014crisislex}. One group of researchers used text mining and statistical approaches to understand crises \cite{karami2020twitter, zou2018mining}, another group of researchers focused on clustering text data to identify a group of tweets that belong to disaster \cite{ashktorab2014tweedr, olteanu2014crisislex}. Later, different traditional machine learning models are used to analyze Twitter data and predict disaster or emergency situations where words of a tweet are represented as embeddings \cite{palshikar2018weakly, 
algur2021classification, singh2019event}. For example, Palshikar et al. \cite{palshikar2018weakly} proposed a weakly supervised model where words are presented with a bag of words (BOW) model. Moreover, frequency-based word representation is used in \cite{algur2021classification} for disaster prediction from Twitter data using Naive Bayes, Logistic Regression, Random Forest, and SVM methods. The authors in \cite{singh2019event} used a Markov-based model to predict the location of Tweets during a disaster. In a recent work \cite{pota2021multilingual}, the authors proposed a pre-processing method for BERT-based sentiment analysis of tweets. However, it is interesting to explore the model performance on different word embeddings to observe how the context words help to predict a tweet as a disaster.


\section{Methodology}
In this section, we discuss our approach of leveraging word embedding for disaster prediction from Twitter data using machine learning methods. We consider three types of word embeddings, 1) bag of words (BOW), 2) context-free, and 3) contextual embeddings. The word embeddings are used in both traditional machine learning methods and deep learning models as input for disaster prediction.

\subsection{BOW embeddings} 
The bag-of-words (BOW) model is a common approach for text representation of a word document. If there are $V$ words in a text vocabulary, then BOW is a binary vector or array of length $|V|$ where each index of the array is used to present one word of the vocabulary. If a word exists in a document, then the corresponding array index of the word becomes one; otherwise, it contains zero. We use BOW embeddings of Twitter data in three traditional machine learning methods such as decision tree, random forest, and logistic regression to predict the sentiment of a tweet. 

Even though BOW is good for representing words of a document, it loses contextual information because the order of words is not recorded in the binary structure. However, contextual information is required to understand and analyze the sentiment of a text. For this reason, we also plan to use context-based embeddings for this sentiment analysis task. 

\subsection{Context-free embeddings} 
Many existing research works proposed to learn word embeddings based on the co-occurrences of word pairs in documents. GloVe \cite{pennington2014glove} is one common method for learning word embeddings from the co-occurrences of words in documents. However, neural network-based models such as Skip-gram \cite{DBLP:journals/corr/abs-1301-3781}, FastText \cite{bojanowski2016enriching} became popular recently to learn word representations from documents and used for sentiment analysis.

In our research study, we use the pre-trained embeddings of three context-free embedding models (GloVe, Skip-gram, FastText) in a neural network-based model to analyze the sentiment of tweet data and predict disaster types tweets. To represent a tweet in context-free embeddings, we took the average of word embeddings of a tweet following the same strategy of \cite{kenter2016siamese}. For the calculated vector of a tweet, we use softmax to predict the sentiment of the tweet. Let's suppose that the vector of a tweet is $v$, we have a set of labels, $L$ = $\{$"positive", "negative"$\}$  and $Z$ $\in$ $\mathbb{R}^{(|L| \times d)}$  is the weight matrix of softmax function. Then, the  probability of the tweet to be positive or disaster is calculated as follows

\begin{equation*}
p(y(l_i)=1)=\frac{e^{Z_i \cdot v}}{\sum_{l_k \in L}e^{Z_k \cdot v}}
\end{equation*}

Recently, deep neural networks are also used for sentiment analysis. To observe how the context-free embeddings work on deep neural networks, we used a bidirectional recurrent neural network with LSTM gates \cite{hochreiter1997long}. The Bi-LSTM model processes the input words of a tweet from right to left and in reverse. The Bi-LSTM block is followed by a fully connected layer and sigmoid function to output as an activation function.

\subsection{Contextual embeddings} 
Unlike the other word embeddings, BERT embeddings \cite{devlin2018bert} generates different vectors for the same word in different contexts. Recent advances in NLP have shown that BERT model have outperformed traditional embeddings in different NLP tasks, like entity extraction, next sentence prediction. In our study, we plan to investigate how well do contextual embeddings work better than traditional embeddings in sentiment analysis. For this purpose, we use the pre-trained embeddings of BERT models in the same neural network models to predict disaster types tweets. 

\begin{table*}
\centering
\caption{Sample pre-processed tweets}
\begin{tabular}{l|l} 
\hline
\textbf{Tweet (original)} & \textbf{Tweet (after preprocessing)}\\ \hline
 $\#$RockyFire Update =$>$ California Hwy. 20 closed  in both &  rockyfire update    california hwy  20 closed \\
directions due to Lake County fire - $\#$CAfire $\#$wildfires & directions due lake county fire cafire  wildfires \\ \hline

$@$TheAtlantic That or they might be killed in an airplane  & theatlantic might killed airplane accident 
\\
accident in the night a car wreck$!$ & night car wreck  \\ \hline
\end{tabular}
\label{tab:data_processing}
\end{table*}

\begin{table}
\centering
\caption{Training data statistic}
\begin{tabular}{l|l} 
\hline
Total train data & 7,613\\ \hline
Total positive data (or disaster tweets) & 3,271\\ \hline
Total unique words  & 21,940\\ \hline
Total unique words with frequency $>$ 1& 6,816\\ \hline
Avg. length of tweets & 12.5\\ \hline
Median length of tweets & 13\\ \hline
Maximum length of tweets & 29\\ \hline
Minimum length of tweets & 1\\ \hline
\end{tabular}
\label{tab:data_statistic}
\end{table}

\begin{figure}[h!]
	\centering
	\includegraphics[scale=0.5]{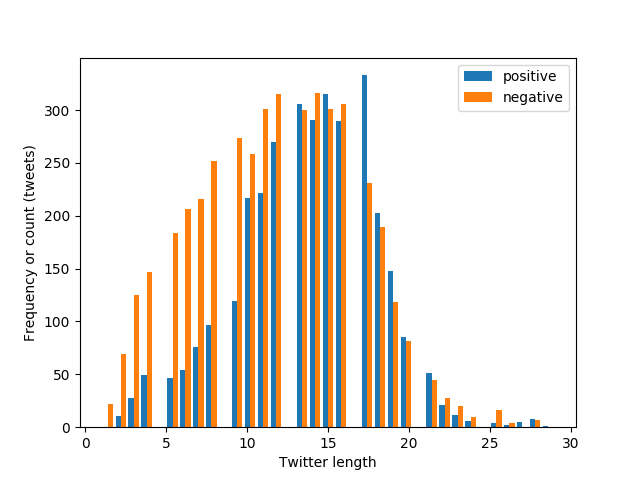}
	\caption{Twitter length distribution in training data}
	\label{fig:word_distribution}
\end{figure}

\section{Dataset}
For this study, we used a Twitter dataset from a recent Kaggle competition (Natural Language Processing with Disaster Tweets \footnote{\label{data_label} https://www.kaggle.com/c/nlp-getting-started}). Kaggle competition is a very well-known platform for machine learning researchers where many research agencies share their data to solve different types of research problems. For example, many researchers used data from Kaggle competitions to analyze real-life problems and propose models to solve the problems, such as sentiment analysis, feature detection, diagnosis prediction \cite{koumpouri2015evaluation, tolkachev2020deep, iglovikov2017satellite, yang2018deep, yang2020computational}.

In the selected Kaggle competition, a dataset of 10,876 tweets is given to predict which tweets are about real disasters and which ones aren't using machine learning model. This dataset has two separate files, train (7,613 tweets) and test (3,263 tweets) data, where each row of the train data contains id, natural language text or tweet, and label. The labels are manually annotated by humans. They labeled a tweet as positive or one if it is about real disaster, otherwise as negative or zero. On the other hand, the test data has D and natural language text but no label. The competition site stores the labels of test data privately and uses that to calculate test scores based on user's machine learning model predictions and create leader-board for the competition based on the test score. Moreover, this dataset was created by the figure-eight company and originally shared on their website \footnote{\label{data_source} https://appen.com/open-source-datasets/}. 

We used the training data to train different machine learning models and predict test data labels using trained models. We reported both the train and test data score in our experiment. Note that our purpose is not to get a high score in the competition, rather use Twitter data to study our research goals. 

\subsection{Data pre-processing} Since the twitter data is natural language text, and it contains different types of typos, punctuation, abbreviations, and numbers. For this reason, before training machine learning models on the natural language text, a text pre-processing step is required to remove stop words and word tokenization. Hence, we removed all the stopwords and punctuations from the training data and converted all the words into lower-case letters. Table \ref{tab:data_processing} shows some pre-processed tweets with the original tweets.

\subsection{Data analysis} Before running any machine learning methods on our data, we analyzed our dataset to obtain some insights about the data. Table \ref{tab:data_statistic} shows some statistical results on the training data after pre-processing the text. From the table, we find that there are 43$\%$ tweets that are annotated as real disasters and 57$\%$ are not. There are a total of 21,940 unique words, while only 6,816 words have frequency $>$1. The average length of tweets is 12.5. However, it is important to check the length of positive and negative tweets separately to verify whether they have common characteristics. Figure \ref{fig:word_distribution} shows word distribution for both the positive and negative tweets. The figure shows many negative tweets with small word lengths ($<$10), but most positive and negative tweets are in a word length of 10 to 20.

\begin{figure*}[h!]
	\centering
	\includegraphics[scale=0.4]{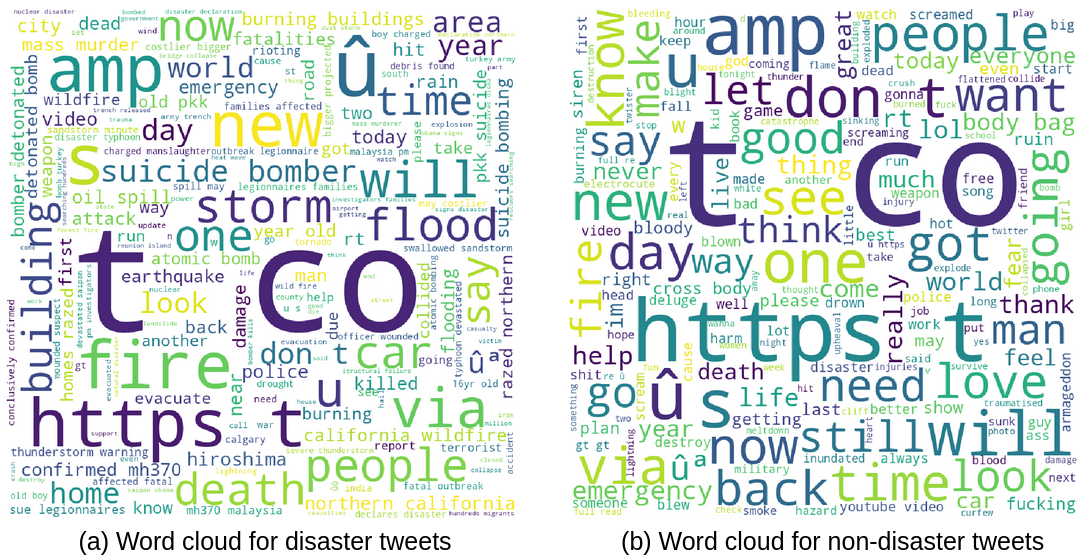}
	\caption{Showing the most frequent words in training data}
	\label{fig:word_cloud}
\end{figure*}

We also analyzed the word frequency for positive and negative tweets. Figure \ref{fig:word_cloud} shows the most frequent words in a word cloud where the high font of a word presents high frequency. We can find some common words in both types of tweets (i.e., https, t, co, people). However, Figure \ref{fig:word_cloud}(a) highlights many disaster-related words like storm, fire, bomber, death, and earthquake. On the other hand, Figure \ref{fig:word_cloud}(b) highlights daily used words such as think, good, love, now, time. From this figure, it is clear that the most frequent words are different in the two types of tweets, and understanding the meaning of words is important to classify them.

\section{Experiments}
In our experimental study, we conduct several experiments based on the real Twitter data to predict disaster-types tweets. At first, we describe the experimental settings and model training procedures in this section. Then, we analyze the experimental results in detail.


\subsection{Experimental settings} 

\subsubsection{Traditional ML models with BOW embeddings} 
From the Table \ref{tab:data_statistic}, we find that the training data has 21,940 unique words where 6,816 words have frequency more than 1. To avoid infrequent words, we considered only the vocabulary of 6,816 words in our BOW representations. To represent a tweet in BOW embeddings, we took a binary array of 6,816 length where it had 1 if a word of tweet was present in the vocabulary, otherwise 0.
We used the BOW embeddings to predict the sentiment of a tweet using three traditional machine learning models, 1) decision tree, 2) random forest and 3) logistic regression. We used python Sklearn package \footnote{https://scikit-learn.org/stable/} and used all the default parameters to train the models on our train dataset. After training the model, we used the test data to get labels and submit that in Kaggle to have test score.

\subsubsection{Deep learning models with context-free embeddings} 
For this experiment, we chose three context-free methods, 1) Skip-gram \cite{DBLP:journals/corr/abs-1301-3781}, 2) FastText \cite{bojanowski2016enriching}, and  2) GloVe \cite{pennington2014glove}. We used publicly available pre-trained embeddings of Skip-gram and fastText models that are trained on Wikipedia data \footnote{ https://nlp.stanford.edu/projects/glove/}. The pre-trained embeddings of FastText is collected from Mikolov et. al. 2018 \cite{mikolov2018advances}. The size of all the pre-trained embeddings or feature is 300. 

The proposed softmax model is trained for 100 epochs using a stochastic gradient algorithm to minimize the categorical cross entropy loss function. We took 1$\%$ of training data as validation data and used the validation data to stop the training model if the loss value for the validation data didn't decrease in last ten epochs. Similarly to the softmax model, we also trained our Bi-LSTM model using batch gradient descent algorithm for 100 epochs to minimize the binary cross entropy loss function. We followed the same stop rule for this model. 
\subsubsection{Deep learning models with contextual embeddings} 
To obtain contextual embeddings, we downloaded publicly available pre-trained BERT model  (Bert-base-uncased) \cite{devlin2018bert} from the official site of the authors \footnote{https://github.com/google-research/bert}. We gave tweets as inputs in the BERT model and took the hidden states of the [$CLS$] token of the last layer from the model as embeddings of the given tweets. Then, the embeddings is used in our sigmoid model to predict sentiment of tweets. The same setting is used in a previous paper \cite{ji2021does} to predict patient diagnosis from medical note words using pre-trained BERT model.

Moreover, we can find embeddings of each words of a tweet from the pre-trained BERT model. The BERT's pre-trained word embeddings are used as input to our Bi-LSTM model. The authors of \cite{lu2020vgcn} used the similar setting for the sentiment analysis of text data.

\subsection{Evaluation metric}
Three different metrics are used in our experiment to evaluate the performance of the machine learning models on the disaster prediction task such as, 1) accuracy, 2) F1 score, and 3) Area Under the Curve (AUC). In our experiment, we considered disaster tweets as 'positive class' and others as 'negative class'. Hence, True Positive (TP) means the actual disaster tweets that are predicted as disaster while False Positive (FP) shows the tweets that are actually false, but predicted as true. True Negative (TN) and False Negative (FN) imply in the same way. The accuracy is the number of correctly predicted tweets among all of the tweets and it is calculated as follows. 
\begin{equation*}
\text{Accuracy (Acc)} =\frac{\text{TP+TN}}{\text{ TP+FP+TN+FN} }
\end{equation*}

 F1 score is another popular metric to test predictive performance of a model. The F1 score is measured by the harmonic mean of recall and precision where recall means the number of true labels are predicted by a model among the total number of existing true labels and precision means the number of true labels are predicted by a model divided by the total number of labels are predicted by the model. The F1 score is calculated as follows.

\begin{equation*}
\text{Recall (R)} =\frac{\text{TP}}{\text{ TP+FN} }
\end{equation*}

\begin{equation*}
\text{Precision (P)} =\frac{\text{TP}}{\text{ TP+FP} }
\end{equation*}

\begin{equation*}
\text{F1 score (F1)} =\frac { \text{2 $\times$ (P $\times$ R)} }{ \text{(P+R)} }
\end{equation*}

On the other hand, AUC tells us how much a model is capable of distinguishing between classes. The higher score of the AUC means the model works better at predicting negative classes as zero and positive classes as one. 

\subsection{Experimental results} 
\subsubsection{Quantitative results} 
Table \ref{tab:prediction_results} provides the results of all the machine learning models on the disaster prediction tasks for all three types of embeddings. The table shows results for both the training and test data. Since the test data results are collected from the Kaggle competition, we only can report the accuracy score. 

The table shows that the logistic regression model has the best results for the BOW embeddings among the three traditional machine learning models. However, the results of neural network models for context-free embeddings are better than the traditional machine learning models that used context-free embeddings as inputs. Among the three context-free embeddings (Skip-gram, FastText, GloVe), the GloVe with Bi-LSTM model has the best train and test score for all the three evaluation metrics. Note that the results also show us that deep learning model like Bi-LSTM has better results than the shallow neural network model such as the softmax model.

Moreover, when we used the same shallow neural network and deep learning models for contextual embeddings such as BERT; we found that there are 2$\%$ improvements on AUC and Acc over the context-free embeddings. It means that contextual embeddings are helpful and have the best performance for the disaster prediction task.

\begin{table}
\centering
\caption{Performance of different machine learning models on disaster prediction for different types of word representations or embeddings}
\begin{tabular}{|l|l|l|l|l|} 
\hline
\textbf{Model}  & \multicolumn{3}{l|}{ \textbf{Train data} } & \textbf{Test data}     \\ 
\hline
   & AUC &F1 & Acc  & Acc \\ 
\hline
\multicolumn{5}{|l|} { \textbf{BOW embeddings} } \\ \hline
Decision tree          &0.6320 &0.5896 &0.6273 &0.6380\\ \hline
Random forest          &0.8313 &0.7320 &0.7848 &0.7042\\ \hline 
Logistic regression    &\textbf{0.8660} &\textbf{0.7443} &\textbf{0.7927} &\textbf{0.7293}\\ \hline \hline
\multicolumn{5}{|l|} { \textbf{Context-free embeddings}} \\ \hline
Skip-gram+Softmax     &0.8281 &0.7301 &0.7769  &0.7649\\ \hline
FastText+Softmax      &0.8336 &0.7231 &0.7769  &0.7826\\ \hline
GloVe+Softmax         &0.8246 &0.7323 &0.7717  &0.7827\\ \hline

Skip-gram+Bi-LSTM      &0.8272 &0.7440 &0.7808  &0.7775\\ \hline
FastText+Bi-LSTM      &0.8327 &0.7369 &0.7817  &0.7955\\ \hline
GloVe+Bi-LSTM         &\textbf{0.8351}  &\textbf{0.7500} &\textbf{0.7991} &\textbf{0.8093}\\ \hline \hline

\multicolumn{5}{|l|} { \textbf{Contextual embeddings} }\\ \hline
BERT+Softmax          &0.8513  &0.8254 &0.8292 &0.8250\\ \hline 
BERT+Bi-LSTM          &\textbf{0.8578}  &\textbf{0.8316} &\textbf{0.8351} &\textbf{0.8308}\\ \hline

\end{tabular}
\label{tab:prediction_results}
\end{table}

\subsubsection{Qualitative results} 
Table \ref{tab:prediction_results} shows us quantitative results for the prediction of disaster tweets where the neural model with contextual embeddings outperformed the other models. However, it is difficult to understand from the result that when the contextual embeddings predict a disaster tweet successfully while context-free models fail. For this purpose, we observe the prediction results of the Bi-LSTM model for both the context-free (GloVe) and contextual embeddings (BERT). Table \ref{tab:sample_prediction} shows the model predictions with true labels for some sample tweets. From the table, we can find that the predictions for GloVe embeddings for the first two tweets are positive, maybe because of the word, ``$accident$'', in the tweets, but the true labels are negative for the two tweets. If we read the tweets, then we can understand that the tweets are not related to disaster or crisis. On the other hand, since the BERT model generates word embeddings based on the context words, it successfully predicts the tweets as negative.

The predictions for GloVe embeddings for the third and fourth tweets are false while they are true. Note that no disaster-related words are used in the two tweets, but the tweets described serious situations. The predictions for BERT embeddings are also correct for the third and fourth tweets. The predictions of GloVe and BERT embeddings for the fifth and sixth tweets of Table \ref{tab:prediction_results} are correct. Since there are some disaster-related words (i.e., suicide, bomber, bombing) in the tweets, both models successfully labeled them.

After analyzing the results of Table \ref{tab:sample_prediction}, it can be implied that the context-free embeddings are helpful to predict a tweet as a disaster if disaster-related words (i.e., accident, bomb) exist in the tweet. In contrast, contextual embeddings help to understand the context of a tweet that is challenging and important for the sentiment analysis task. Although every tweet has a short length text, contextual embeddings works efficiently to understand the sentiment of a tweet.

\begin{table*}
\centering
\caption{Showing sentiment predictions of Bi-LSTM model for pre-trained GloVe and BERT embeddings}
\begin{tabular}{l|l|l|l|l} 
\hline
  & \textbf{Sample tweets} & \multicolumn{2}{l|}{ \textbf{Prediction} } & \textbf{True}\\
  & &\textbf{GloVe} & \textbf{BERT} & \textbf{label} \\  \hline

1 & I swear someone needs to take it away from me, cuase I'm just accident prone. & Yes & No &No\\ \hline
2 &$@$Dave if I say that I met her by accident this week- would you be& & & \\ 
  &super jelly Dave? :p & Yes & No &No\\ \hline
3 &Schoolgirl attacked in Seaton Delaval park by 'pack of animals' & No & Yes &Yes\\ \hline
4 &Not sure how these fire-workers rush into burning buildings  & &  & \\ 
  &but I'm grateful they do. $\#$TrueHeroes & No & Yes &Yes\\ \hline
5 & A suicide bomber has blown himself up at a mosque in the south & Yes & Yes &Yes\\ \hline
6 &Bombing of Hiroshima 1945 & Yes & Yes &Yes\\ \hline

\end{tabular}
\label{tab:sample_prediction}
\end{table*}

\section{Conclusion}
In this paper, we described an extensive analysis for predicting disaster from Twitter data using different types of word embeddings. Our experimental results show that contextual embeddings have the best result for predicting disaster from tweets. We also showed that deep neural network models outperformed traditional machine learning methods in the disaster prediction task. Advance deep neural network models such as multi-layer convolutional models can also be used for this prediction task to achieve higher accuracy.

\bibliographystyle{ACM-Reference-Format}
\bibliography{sample-base}

\clearpage  

\end{document}